\documentclass[]{article}

\usepackage{amssymb}
\usepackage{amsmath}
\usepackage{physics}
\usepackage[dvipsnames]{xcolor} 
\usepackage{lineno}
\usepackage{float} 
\usepackage{lscape} 
\usepackage{rotating}
\usepackage{longtable} 
\usepackage{lineno}
\modulolinenumbers[5]

\usepackage{booktabs}
\usepackage{xcolor}
\usepackage{tabularx}
\usepackage{array}
\usepackage{multirow}
\usepackage{siunitx}
\usepackage{authblk}
\title{Accuracy-Efficiency Trade-Offs in Spiking Neural Networks: A Lempel-Ziv Complexity Perspective on Learning Rules}

\author{Zofia Rudnicka}
\author{Janusz Szczepanski}
\author{Agnieszka Pregowska}
\affil{Institute of Fundamental Technological Research, Polish Academy of Sciences,	Pawinskiego~5B, 02--106 Warsaw, Poland}

\date{} 

\begin{document}
		\maketitle

\begin{abstract}
	Training spiking neural networks (SNNs) remains challenging due to temporal dynamics, non-differentiability of spike events, and sparse event-driven activations. This paper studies how the choice of learning paradigm (unsupervised, supervised, and hybrid) affects classification performance and computational cost in temporal pattern recognition.
	Building on our earlier study \cite{Rudnicka2026}, we use Lempel-Ziv complexity (LZC) as a compact, decision-relevant descriptor of spike-train temporal organization to quantify how different learning rules reshape class-conditional temporal structure. The pipeline combines a leaky integrate-and-fire (LIF) SNN with an LZC-based decision rule. We evaluate learning rules on synthetic sources with controlled temporal statistics (Bernoulli, two-state Markov, and Poisson spike processes) and on two-class subsets of MNIST and N-MNIST.
	Across datasets, gradient-based learning achieves the highest accuracy but at high computational cost, whereas bio-inspired rules (e.g., Tempotron and SpikeProp) offer favorable accuracy--efficiency trade-offs. These results highlight that selecting a learning rule should be guided by application constraints and the desired balance between separability and computational overhead.
\end{abstract}



\section*{keyword}
	spiking neural networks, learning rules, spike trains, temporal classification, Lempel-Ziv complexity, computational efficiency

\section{Introduction}
Backpropagation is a gradient-based learning rule commonly applied in networks training \cite{Lee2016}. Artificial Neural Networks (ANNs) and Spiking Neural Networks (SNNs) are both inspired by biological neural systems, however, they differ from the computation point of view. While ANNs dominate deep learning due to their simplicity, SNNs better mimic real neurons by using spike-based communication. However, training SNNs to match ANN accuracy, especially with backpropagation-based methods \cite{Lee2016}, requires significant hardware resources and is computationally expensive \cite{Zhang2026}. As a consequence, SNNs training remains a challenge due to their complex dynamics and the non-differentiable nature of spike-based activity \cite{Rudnicka2024,Zhao2025}. 
\par
Learning rules can be categorized into three main types: unsupervised learning (for example, Hebbian learning \cite{Hebb2002}, Symmetric Spike-Driven Synaptic Plasticity (SDSP), and Spike Timing Dependent Plasticity (STDP) \cite{Khoee2024}, supervised learning (like backpropagation (BP) \cite{Lee2016}, Spike-Timing-Dependent Backpropagation (STBP) \cite{Tang2023}, tempotron \cite{Gutig2006,Chen2025}, SpikeProp \cite{Shrestha2015}, its direct application to LIF  \cite{Laddach2024} Chronotron \cite{Florian2010}, and Remote Supervised Method (ReSuMe \cite{Song2024}), and hybrid learning (including Artificial-to-Spiking Neural Network conversion (ANN-SNN conversion)) \cite{Midya2019}, and Reward-based STDP \cite{Khoee2024}.
\par
A commonly used training method in SNNs is SpikeProp, which is an arithmetic-based learning algorithm conceptually similar to the  BP algorithm in ANNs \cite{Rajagopal2023}. However, arithmetic-based learning rules may be not a right choice for building biologically efficient networks. To achieve more biologically plausible learning, several alternative methods have been developed. For example, STDP that is inspired by biological synaptic plasticity, where neurons strengthen or weaken connections based on temporal contiguity. However, despite its efficiency in synaptic adaptation, STDP does not match the classification performance of BP-based ANN training \cite{Lee2016}. Moreover, spatio-temporal backpropagation (STBP) extends the traditional backpropagation algorithm to handle both spatial and temporal dependencies in spiking neural networks. This makes it possible to train deep SNNs using gradient-based optimization \cite{Wu2025}. Other commonly used learning algorithms in SNN training are ReSuMe \cite{Song2024,Guo2025}, and Chronotron \cite{Florian2010}. They are effective in adaptation to structured spike patterns, however, both are computationally demanding and require significant training time. An interesting SNN learning rule, namely tempotron was proposed by \cite{Gutig2006}. It is particularly effective for temporal coding, where information is embedded in the precise timing of spikes \cite{Pregowska2016}. In the study \cite{Patankar2025} training is conducted using the tempotron learning rule, while optimization is achieved through STDP. This combination significantly improves processing speed and reduces computational complexity compared to one can say "conventional" spiking neural network methods. In turn, \cite{Wu2019} proposed a neuron normalization technique and an explicitly iterative neuron model, which resulted in a significant increase in the SNNs' learning rate. 
Recently, the Spiking Gate (SG) ResNet architecture with an attention spike decoder (ASD) has been proposed to address gradient vanishing and improve spike decoding, achieving state-of-the-art performance on several benchmarks and enabling the first direct-training hybrid SNN-ANN detector \cite{Zhang2023a}. Another training strategy involves first training an ANN (typically using perceptrons) and then converting it into an SNN with an equivalent structure \cite{Midya2019}. This conversion leverages the strengths of ANN training while allowing SNNs to benefit from energy-efficient spike-based processing. However, a major drawback is that this approach fails to capture temporal spike dynamics, as reliable frequency estimation requires a non-trivial time period. On the other hand, to optimize data efficiency, various active learning approaches have been introduced, including Bio-Inspired Active Learning (BAL), Bio-Inspired Active Learning based on Firing Rate (BAL-FR), and Bio-Inspired Active Learning utilizing Membrane Potential (BAL-M) \cite{Zhang2023}. These methods identify and select training samples by analyzing the discrepancies between empirical observations and generalization behavior, ensuring that previously unrecognized patterns are incorporated into the learning process. In turn, \cite{Tang2023} proposes the AC2AS framework, which couples activation consistency between ANN and SNN to enable fast and memory-efficient training of spiking neural networks.
\par

We use a previously introduced LZC-SNN analysis framework \cite{LempelZiv1976} to study how learning rules reshape temporal spike-train organization. In this work, LZC is treated as a compact descriptor of temporal novelty and regularity, and the downstream classification performance is interpreted as a probe of separability induced by learning rather than as evidence of a universally strong stand-alone classifier.
It turned out that the capability of bio-inspired neural networks in biosignal classification is highlighted, emphasizing their high precision, sensitivity, and reliability in distinguishing physiological patterns.

This study builds explicitly upon our earlier work \cite{Rudnicka2026}, in which Lempel--Ziv complexity (LZC) was introduced as an analysis tool to investigate how different neuron models affect spiking neural network dynamics and downstream separability. In contrast to that work, the present manuscript shifts the scientific focus from neuron-model selection to the role of learning algorithms themselves.

Rather than re-introducing the coupling between LZC and spiking neural networks, we use LZC here as a compact temporal descriptor to quantify how unsupervised, supervised, and hybrid learning paradigms reshape the temporal organization of spike trains. This shift allows us to analyze learning rules not only in terms of classification accuracy, but also in terms of how they reorganize temporal variability, correlations, and computational efficiency under controlled input statistics.

While the underlying SNN architecture and part of the synthetic data generation are shared between the two studies, the scope and objectives differ substantially. Here, LZC is not used merely as a descriptive post hoc metric but as a decision-relevant representation to analyze the effect of unsupervised, supervised, and hybrid learning rules on class-conditional temporal complexity, robustness, and computational cost.

In this study, we address the problem of comparing learning rules in spiking neural networks from the perspective of temporal organization and computational efficiency. We propose a unified experimental protocol covering unsupervised, supervised, and hybrid learning paradigms under controlled temporal statistics, complemented by two-class MNIST and N-MNIST subsets. Within this framework, we demonstrate that learning rules with similar accuracy can produce substantially different class-conditional Lempel-Ziv complexity profiles, highlighting differences in temporal structure that are not reflected in accuracy metrics alone. By jointly analyzing accuracy and computational overhead, we identify operating regimes in which biologically inspired learning rules provide favorable efficiency-accuracy trade-offs. We also discuss the limitations of scalar complexity-based decisions, particularly under class imbalance and overlapping complexity distributions, and position Lempel-Ziv complexity as a decision-relevant descriptor rather than a universal classifier.

\subsection{Relation to Prior Work}
\label{sec:relation}

Table~\ref{tab:comparison_prior} summarizes the key differences between the present study and our earlier study \cite{Rudnicka2026}. Although both studies employ a similar SNN architecture and synthetic spike-based datasets, the scientific questions, evaluation criteria, and conclusions are fundamentally different. The earlier study focused on the influence of neuron models on SNN dynamics, whereas the present work isolates the role of learning algorithms and their impact on temporal information structure, robustness, and computational efficiency.

\begin{table*}[t]
\centering
\caption{Comparison between the present study and prior work~\cite{Rudnicka2026}.}
\label{tab:comparison_prior}
\renewcommand{\arraystretch}{1.2}
\begin{tabularx}{\textwidth}
	{>{\raggedright\arraybackslash}p{4cm} X >{\raggedright\arraybackslash}X}
	\toprule
	\textbf{Aspect} 
	& \textbf{Neuroinformatics (2026)~\cite{Rudnicka2026}} 
	& \textbf{This work} \\
	\midrule
	
	Primary focus 
	& Neuron model dynamics 
	& \textbf{Learning rule dynamics} \\
	
	Role of LZC 
	& Descriptive analysis metric 
	& \textbf{Decision-relevant temporal descriptor} \\
	
	Learning algorithms 
	& Fixed learning setup 
	& \textbf{Unsupervised, supervised, and hybrid learning} \\
	
	Evaluation criteria 
	& Accuracy and neuron behavior 
	& \textbf{Accuracy, LZC profiles, and computational cost} \\
	
	Scientific question 
	& How neuron models affect SNNs 
	& \textbf{How learning rules reshape temporal structure} \\
	
	\bottomrule
\end{tabularx}
\end{table*}

To ensure transparency, we note that the network topology and part of the synthetic data generation follow \cite{Rudnicka2026}. However, the present study introduces a different experimental axis (learning rules), new evaluation dimensions (computational cost and complexity profiles under training), and additional datasets to assess generalization under controlled constraints.

Importantly, although both studies share a comparable LIF-based network topology and part of the synthetic data generation, the evaluated object is fundamentally different. The earlier work isolated the impact of neuron models under a fixed learning setup, whereas the present study isolates the impact of learning rules under a fixed neuron model. Consequently, the reported effects concern different mechanisms: here we focus on how learning dynamics reshape temporal correlations and variability in spike trains, as exposed by shifts in class-conditional LZC distributions and accuracy-efficiency trade-offs.

\section{Background on Learning Rules in Spiking Neural Networks}

Rather than treating learning rules solely as optimization procedures, we interpret them as mechanisms that actively shape the temporal structure of spike trains. Since Lempel-Ziv complexity directly quantifies the emergence of novel temporal patterns, different learning paradigms are expected to induce distinct complexity profiles even when classification accuracy is similar. In this work, learning algorithms are therefore grouped according to how they modulate temporal correlations, variability, and information flow in spiking activity, which in turn affects complexity-based pattern recognition.

In this section, we provide the background on the learning rules considered in this study and introduce the notation used throughout the paper. We first summarize the symbols and variables employed in the remainder of the manuscript (Section~\ref{sec:notation}). We then group the learning algorithms for spiking neural networks into three main families: unsupervised, supervised, and hybrid rules. For each family, we briefly recall the underlying learning principle, present the corresponding mathematical formulation, and highlight its main advantages and limitations in the context of temporal pattern recognition. This structured overview prepares the ground for the description of the classification task and the proposed LZC-based SNN classifier in the subsequent sections.

\subsection{Notation}
\label{sec:notation}

In Table \ref{tab:notation} the basic notation is presented.

\begin{table}[H]
\centering
\renewcommand{\arraystretch}{1.1}
\begin{tabularx}{\columnwidth}{>{\raggedright\arraybackslash}p{3.4cm} X}
	\toprule
	\textbf{Notation} & \textbf{Description} \\
	\midrule
	
	$\mathbf{x}=[x_{1},x_{2},\dots,x_{n}] \in \mathbb{R}^{n}$ 
	& Synaptic weight vector \\
	$V_{m}(t) \in \mathbb{R}$ 
	& Membrane potential at time $t$ \\
	$\tau_{m} \in \mathbb{R}^{+}$ 
	& Membrane time constant \\
	$I(t) \in \mathbb{R}$ 
	& Input current at time $t$ \\
	$t_{f} \in \mathbb{R}^{+}$ 
	& Neuron firing time \\
	$S_{i}(t) \in \{0,1\}$ 
	& Binary spike indicator at time $t$ \\
	$t_{i}^{k} \in \mathbb{R}^{+}$ 
	& Time of the $k$-th spike of neuron $i$ \\
	$H(V_{m}(t)) \in \{0,1\}$ 
	& Heaviside step function \\
	$\eta \in \mathbb{R}^{+}$ 
	& Learning rate \\
	$\tau_{+}, \tau_{-} \in \mathbb{R}^{+}$ 
	& STDP time constants \\
	$A \in \mathbb{R}$ 
	& SDSP learning rate \\
	$A_{+}, A_{-} \in \mathbb{R}^{+}$ 
	& STDP amplitudes \\
	$S_{\mathrm{pre}}(t), S_{\mathrm{post}}(t) \in \{0,1\}$ 
	& Pre- and postsynaptic binary spike trains \\
	
	$t_{\mathrm{pre}}, t_{\mathrm{post}} \in \mathbb{R}^{+}$ 
	& Pre- and postsynaptic spike times \\
	
	$K(t): \mathbb{R}\rightarrow\mathbb{R}$ 
	& Synaptic kernel used in spike-timing rules (e.g., Tempotron) \\
	
	$r(t) \in \mathbb{R}$ 
	& Reward signal in reward-modulated plasticity \\
	
	$w_{\mathrm{SNN}}, w_{\mathrm{ANN}} \in \mathbb{R}$ 
	& Synaptic weight in SNN / ANN (conversion setting) \\
	
	$\tau_{\mathrm{syn}} \in \mathbb{R}^{+}$ 
	& Synaptic time constant \\
	
	$U(w_i): \mathbb{R}\rightarrow\mathbb{R}$ 
	& Uncertainty function used in active learning updates \\
	
	\bottomrule
\end{tabularx}
\caption{Basic notation used in this study.}
\label{tab:notation}
\end{table}

Unsupervised learning algorithms learn patterns and structures from unlabeled data without explicit supervision or target output, i.e., provide local synaptic modifications without explicit error signals. Unsupervised learning rules in spiking neural networks adjust synaptic weights based solely on the statistics and timing of pre- and postsynaptic activity, without access to explicit target labels. They provide local, biologically plausible synaptic updates and are particularly suitable for capturing regularities in continuous, unlabeled spike streams. In this work we focus on three widely used unsupervised mechanisms, Hebbian learning, spike-timing-dependent plasticity, and symmetric spike-driven synaptic plasticity, as they form a representative set of bio-inspired rules against which we later compare supervised and hybrid approaches. From an information-theoretic perspective, correlation-driven unsupervised rules tend to suppress temporal variability by reinforcing frequently co-occurring spike patterns. As a consequence, they often lead to reduced LZC values and narrower complexity distributions, which can limit discriminative power for inputs with overlapping temporal statistics.

\subsection{Unsupervised Learning Algorithms}
\label{sec:unp}

Hebbian Learning Rule strengthens synapses when both neurons fire together \cite{Hebb2002}
\begin{equation}
\Delta w_{i} = \eta (S_{pre}(t)S_{post}(t)) .
\label{Hebb}
\end{equation}
A major drawback of Hebbian learning is its decreasing efficiency as the number of hidden layers increases \cite{Amato2019}. Although it remains competitive with up to four layers, its performance diminishes in deeper networks, limiting its scalability for complex architectures.

\subsubsection{Spike Timing Dependent Plasticity} 
Symmetric spike-driven plasticity refers to a class of local learning rules with symmetric weight updates driven by spike events, rather than a single canonical algorithm
\begin{equation}
\Delta w_{i} =
\begin{cases}
	A_{+}e^{-\frac{(t_{post}-t_{pre})}{\tau_{+}}} & t_{post} > t_{pre} \\
	-A_{-}e^{-\frac{(t_{post}-t_{pre})}{\tau_{-}}} & t_{pre} > t_{post} .
\end{cases}
\end{equation}
It is a form of Hebbian learning (\ref{Hebb}). It is essential for modeling the temporal dynamics of biological neural systems. Unlike traditional learning methods, its event-driven nature enables unsupervised learning, allowing neural networks to adapt autonomously to temporal patterns in input data. This makes it valuable for processing spatio-temporal signals, such as sensory data. In addition, its computational demands can be high in large-scale simulations.

\subsubsection{Symmetric Spike-Driven Spike-Based Plasticity}
SDSP is a variant of STDP, however, with symmetric weight updates

\begin{equation}
\Delta w_{i} = A (S_{pre}(t)-S_{post}(t)) .
\end{equation}

It is a biologically plausible learning mechanism that mirrors neural dynamics by focusing on individual spike events. This event-driven approach enables rapid adaptability, making it valuable for real-time learning. However, its sensitivity to single spikes poses challenges in noisy environments, requiring advanced filtering techniques.

\subsection{Supervised Learning Algorithms}
Supervised learning maps input data to output labels based on labeled dataset. It relies on error-based updates. They adjust synaptic weights based on an explicit error or teacher signal that encodes the desired network output. In contrast to unsupervised rules, they are designed to optimize task-specific performance, such as classification accuracy, at the cost of reduced biological plausibility and typically higher computational complexity. Here we review representative supervised learning rules for SNNs, including gradient-based methods (backpropagation and STBP) and spike-timing-based algorithms (tempotron, SpikeProp, Chronotron, and ReSuMe), all of which are later evaluated within the same classification framework. Error-driven learning rules explicitly optimize class separability, which manifests not only in improved accuracy but also in increased divergence between class-conditional LZC distributions. This effect highlights that supervised learning influences not only decision boundaries but also the internal temporal organization of spike trains.

\subsubsection{Backpropagation Algorithm}

Backpropagation is a gradient-based learning rule commonly applied in networks training \cite{Lee2016}.

\begin{equation}
\Delta w_{i} = -\eta \frac{\partial E}{\partial w_{i}} \label{BP}
\end{equation}
where \( E \) is the error function. The weight update follows a chain rule
\begin{equation}
\frac{\partial E}{\partial w_{i}} = \frac{\partial E}{\partial V_{m}} \cdot \frac{\partial V_{m}}{\partial w_{i}} .
\end{equation}

\subsubsection{Spike-Timing-Dependent Backpropagation}
Spike-Timing-Dependent Backpropagation extends the conventional gradient-based learning rule of backpropagation to spiking neural networks, incorporating spike-timing-dependent plasticity  principles to optimize synaptic weights \cite{Tang2023}. The weight update rule is given by Equation (\ref{BP}). In STBP, the gradient computation follows the chain rule:
\begin{equation} \frac{\partial E}{\partial w_{i}} = \frac{\partial E}{\partial t_{\text{spike}}} \cdot \frac{\partial t_{\text{spike}}}{\partial V_{m}} \cdot \frac{\partial V_{m}}{\partial w_{i}}. \end{equation}

Unlike traditional backpropagation, STBP accounts for the temporal dynamics of spikes, leveraging precise spike timing to adjust synaptic strengths. This biologically inspired approach enables more efficient credit assignment in spiking networks while maintaining compatibility with gradient-based optimization methods.

\subsubsection{Tempotron Learning Rule}
Tempotron learning rule adjust synaptics weight to classify temporal spike patterns \cite{Gutig2006,Yu2014}

\begin{equation}
\Delta w_{i} = \eta \sum_{t \in T_{pre}}K(t_{f}-t)
\end{equation}

The tempotron learning principle is a biologically inspired approach that refines synaptic weights based on the precise timing of incoming spikes rather than just their frequency. Unlike conventional neural models that prioritize weight adjustments or network structures, the tempotron underlines the importance of timing, showing that the "when" of a neural event can be as crucial as the "where" or "how often".

\subsubsection{SpikeProp Algorithm}
SpikeProp applies gradient descent to minimize spike timing errors \cite{Shrestha2015}:
\begin{equation}
\frac{\partial E}{\partial w_{i}} = \sum_{t_{post}} \frac{\partial E}{\partial t_{i}^{actual}} \cdot \frac{\partial t_{i}^{actual}}{\partial w_{i}} .
\end{equation}
SpikeProp extends backpropagation to spike-based networks, enabling gradient-based optimization. It is computationally expensive due to temporal dynamics. Also, requires careful handling of non-differentiability in spike generation.

\subsubsection{Chronotron Learning Rule}

The Chronotron minimizes the actual and target spike times \cite{Florian2010}

\begin{equation}
E = \sum_i \sum_k \frac{(t_{i}^{actual} - t_{i}^{target})^2}{2}
\end{equation}
\begin{equation}
\Delta w_{ij} = -\eta \sum_k (t_{i}^{actual} - t_{i}^{target}) \frac{\partial t_{i}^{actual}}{\partial w_{ij}}
\end{equation}

A key advantage of this approach is its consistent coding for both inputs and outputs, improving the efficiency of SNNs. However, the granularity of the method can significantly increase computational demands, requiring precise parameter tuning and high-quality temporal data. 

\subsubsection{Remote Supervised Method}

ReSuMe combines STDP with supervised learning using a teacher spike train \cite{Song2024}

\begin{equation}
\Delta w_{ij} = \eta \left[ S_{pre}(t) * S_{teach}(t) - S_{pre}(t) * S_{post}(t) \right]
\end{equation}

ReSuMe enables supervised training while maintaining biological plausibility \cite{Taherkhani2015}.

\subsection{Hybrid Learning Algorithms}
Hybrid algorithms integrate elements of both supervised and unsupervised learning. These schemes combine error-driven optimization with local synaptic plasticity or bridge the gap between rate-based ANNs and event-based SNNs. Such methods aim to leverage the robustness and training efficiency of conventional deep learning while retaining the energy efficiency and temporal precision of spike-based computation. In this study we consider three representative hybrid strategies: ANN-to-SNN conversion, reward-modulated STDP, and bio-inspired active learning. Hybrid and reward-modulated learning strategies introduce controlled stochasticity through exploration mechanisms. This property allows them to maintain higher temporal variability under noisy conditions, preventing premature collapse of spike complexity and yielding more stable LZC-based representations for stochastic inputs.

\subsubsection{Artificial-to-Spiking Neural Network conversion} ANN-SNN conversion maps pre-trained ANN weights to SNN synapses \cite{Midya2019}

\begin{equation}
w_{SNN}=\frac{w_{ANN}}{\tau_{syn}} .
\end{equation}
Converting ANNs to SNNs allows the use of existing training methodologies while leveraging SNNs' energy efficiency and biological realism \cite{Yan2025}. This conversion maps ANN activation values to the firing rates of SNN neurons, requiring precise threshold tuning to maintain accuracy. Improper thresholds can lead to excessive or insufficient spiking, affecting performance. 

\subsubsection{Reward-based STDP}

Reward-based STDP modifies STDP with reinforcement learning
\begin{equation}
\Delta w_{ij} = \eta r(t) \left[ S_{pre}(t) * S_{post}(t) \right] .
\end{equation}

\subsubsection{Bio-inspired Active Learning}

Active learning updates weights based on uncertainty \cite{Xie2024}
\begin{equation}
\Delta w_{ij} = \eta \cdot U(w_{i}) \cdot \mathbb{E} \left[ I(S_{post}; S_{pre}) \right] .
\end{equation}

Integrates biological learning mechanisms with active learning strategies to improve training efficiency \cite{Xie2024}. Unlike traditional supervised learning, which relies on large labeled datasets, BAL actively selects the most informative data points for labeling, reducing data requirements while maintaining high performance.

\section{Classification Task}

\begin{figure}
\centering
\includegraphics[width=5.5in]{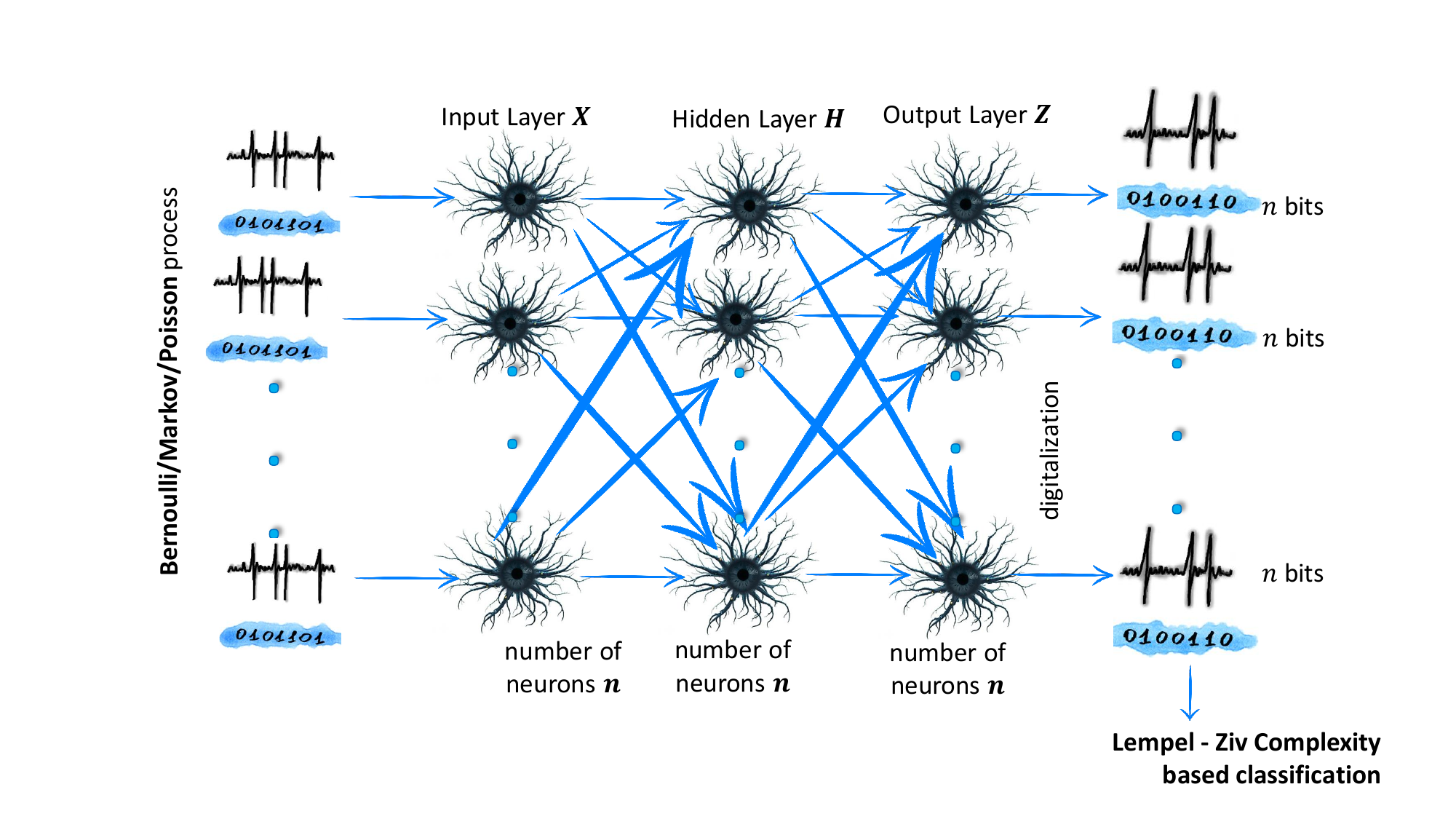}
\caption{The architecture of the considered neural network.}
\label{fig:task}
\end{figure}

The SNN neural network under consideration consists of three layers: input, hidden, and output, each containing $n$ neurons; see Figure \ref{fig:task}. It was based on the Leaky Integrate-and-Fire (LIF) biophysical neuron model that is a fundamental component of SNNs to simulate the subthreshold behavior of the membrane potential of a neuron $U(t) \in \mathbb{R}$ \cite{Dutta2017}. Given an input vector $\mathbf{x}(t) \in \mathbb{R}^{n}$, synaptic weights $\mathbf{w} \in \mathbb{R}^{n}$, and bias $b \in \mathbb{R}$, the net input current is defined as: 
\begin{equation} 
I(t) = \mathbf{w}^{T} \mathbf{x}(t) + b = z(t), 
\end{equation} where $z(t)$ denotes the weighted synaptic drive at time $t$. The membrane voltage evolves according to: \begin{equation} \tau_m \frac{dU(t)}{dt} = -U(t) + z(t), \end{equation} and emits a spike when $U(t) \geq U_{\text{th}}$, after which the voltage is reset to a resting value $U_r$. This model captures the temporal integration and passive leakage of membrane charge in a computationally efficient form.

We considered different sets of neuron parameters and the presented results include the most efficient set of parameters for each learning algorithm. The threshold took values 1.00, 0.50, 0.40, 0.30, 0.20, 0.10, and 0.05. The decline was 0.100, 0.05, 0.03, 0.01, and the learning rates were 0.0500, 0.0100, 0.0010, 0.0098, and 0.0001. We considered neural networks comprising 16, 32, 64, 128, 256, 512, and 1024 LIF neurons per layer. The best of the results obtained are presented in Table \ref{table:results_SNN_unsupervised}, Table \ref{table:results_SNN_supervised}, and Table \ref{table:results_SNN_hybrid}. A study on the effect of neuron count on transmission efficiency in the case of the Levy-Baxter neuron model has been reported in \cite{Paprocki2024}.
\par
The network processes binary sequences of length 1024, which are transformed into $n$-bit spike trains and subsequently converted back into binary sequences. The classification of these output sequences is performed using the Lempel-Ziv complexity (LZC) measure, a well-established metric for evaluating the informational content of a sequence. A binary sequence \( \textbf{x}_n^1 := [x_1, x_2, \dots, x_n] \), where each \( x_i \in \{0,1\}\) the LZ complexity \( C_{\alpha}(\textbf{x}_n^1) \) represents the count of distinct substrings that appear sequentially in \( \textbf{x}_n^1\). The normalized complexity, which assesses the rate at which novel patterns emerge, is defined as:

\begin{equation}
c_{\alpha}(\mathbf{x}_n^1) = \frac{C_{\alpha}(\mathbf{x}_n^1)}{n}\log_{\alpha}{n},
\end{equation}
where  $\alpha = 2 $ for binary sequences.
For purely random sequences, \( c_{\alpha}(\textbf{x}_n^1) \) converges to 1, while for deterministic sequences, it approaches 0. This measure effectively approximates the entropy of ergodic stochastic processes, making it a valuable tool to evaluate randomness and structure within the output of the neural network \cite{LempelZiv1976,Pregowska2019}. For completeness, we performed our analysis for information sources for which the subsequent symbols are more and more correlated \cite{Pregowska2019a}. 

\section{Input Datasets}
In this study, we employed synthetic binary datasets designed to represent different types of temporal dependencies and levels of stochasticity. We consider three types of dataset that consist of Bernoulli sequences (uncorrelated), two-state Markov, and Poisson processes \cite{BritvinaEggermont2006}. Poisson spike trains have been widely used to model neural activity since early studies, owing to their biologically realistic temporal dynamics and the characteristic proportionality between the mean and variance of neural responses, which gives rise to variable interspike intervals \cite{Rieke1997}. Experimental data have confirmed this Poisson property \cite{Rieke1997,BritvinaEggermont2006}. The binary sequence of each process is converted into inputs to LIF models with specified $\Delta t = 1ms$ \cite{Rieke1997}. For the evaluation of the proposed classification task, we applied accuracy the mean squared error (MSE) and \(R^{2}\) score.

Bernoulli sequences are generated from independent Bernoulli processes, where each binary element $b_{i}$ in a given sequence is drawn independently with probability p. For our experiments, we generate two sets of binary sequences $B_{1}$ and $B_{2}$, each sequence of length 1024, using different values of $p$. The independence of events allows us to assess the system’s performance under random binary outcomes with varying probabilities.

Markov sequences are generated from a first-order Markov process, where the probability of each element \(s_{i}\) in a given sequence depends on the previous state \(s_{i - 1}\). For our experiments, two sets of binary sequences of length 1024, \(M_{1} = \{m_{1,k}\}, k = 1, 2, \dots 1000\) and  \(M_{2} = \{m_{2,k}\}, k = 1, 2, \dots 1000\)  are generated with transition probabilities defining the dependency structure. We explore different transition probability configurations to evaluate the impact of state dependencies on classification accuracy.

Poisson sequences are generated from a Poisson process, where events occur independently at a constant average rate $\lambda$. For each Poisson process, spike trains are generated with rate parameters $\lambda_{x}$ and $\lambda_{y}$. These sequences are tested under various rate configurations to observe the effect of spike timing variability on system behavior and classification performance.
This variety of data we employed allows us to systematically assess the strengths and limitations of classical and bio-inspired learning algorithms under different input regimes. Bernoulli sequences test the ability to classify memoryless inputs, Markov sequences require short-term memory handling, and Poisson sequences emulate real-world neural signal variability. Using synthetic data also enables precise control over statistical properties and eliminates noise or artifacts that can confound real-world datasets. This makes it possible to isolate and compare the intrinsic capabilities of the learning rules. Performance was evaluated using classification accuracy, mean squared error (MSE), and $R^{2}$ score.

This synthetic design allows precise control over the temporal structure and stochasticity of inputs, which is critical when evaluating biologically inspired learning rules under known statistical regimes. Unlike benchmark datasets, our generative setup isolates key dynamics relevant to spiking computation.

In addition to the synthetic temporal datasets, we also evaluated the proposed approach on binarized versions of two widely used benchmark datasets: MNIST and N-MNIST, restricted to the digits 0 and 1. The MNIST01 subset consists of grayscale handwritten digits from the original MNIST collection, filtered to include only classes 0 and 1, resulting in a clean two-class pattern-recognition task. Each image was normalized and flattened into a 1,024-dimensional input vector (32 $\times$ 32 cropping/reshaping), consistent with the dimensionality used in our spike-based models. This reduced task serves as a controlled setting to isolate the behavior of the SNN learning rules without confounding factors such as multi-class overlap or high intra-class variability. The N-MNIST01 subset follows the same class restriction but is derived from the neuromorphic N-MNIST dataset, where event-based recordings of moving MNIST digits are captured using a Dynamic Vision Sensor (DVS). Events corresponding to digits 0 and 1 were retained, and the resulting spike streams were temporally discretized into the same fixed-length input representation used for all other experiments. This allows direct comparison between conventional image-based inputs (MNIST01) and event-driven neuromorphic inputs (N-MNIST01) under identical network architectures and training procedures. Input characteristics for the MNIST01 and N-MNIST01 subsets used in our experiments are shown in Table \ref{tab:mnist_stats}. Both MNIST01 and N-MNIST01 contain only the digits 0 and 1, enabling a controlled binary classification setting.
For M-NIST01, images are normalized and converted to a fixed 1,024-dimensional vector. For N-MNIST01, DVS events corresponding to digits 0 and 1 are aggregated into 32 $\times$ 32 bins, producing spike-based representations with identical dimensionality, allowing direct comparison across static and event-driven modalities.

\begin{table}[H]
\centering
\small
\caption{Input characteristics for the MNIST01 and N-MNIST01 subsets used in our experiments.}
\label{tab:mnist_stats}

\setlength{\tabcolsep}{4pt}
\renewcommand{\arraystretch}{1.1}

\begin{tabular}{lccc}
	\hline
	\textbf{Dataset} & \textbf{Samples (train/test)} & \textbf{Input type} & \textbf{Dim.} \\
	\hline
	MNIST01   & 12,665 / 2,115 & Static grayscale images & 1024 \\
	N-MNIST01 & 12,665 / 2,115 & Event-based spike streams & 1024 \\
	\hline
\end{tabular}

\end{table}

The restriction to a binary (0/1) setting is intentional and serves to isolate the effect of learning rules on temporal representations under controlled conditions, rather than to compete with multi-class benchmark performance.

\section{Results and Discussion}
In contrast to conventional evaluations that focus primarily on classification accuracy, our analysis emphasizes how different learning paradigms reshape the temporal information structure of spike trains. Lempel-Ziv complexity is therefore treated as a primary descriptor, while accuracy, MSE, and $R^2$ scores are interpreted as secondary outcomes resulting from these underlying structural changes.

We consider classical and bio-inspired learning algorithms in SNNs. Our prior work \cite{Rudnicka2026} introduced LZC as an analysis tool for assessing the impact of neuron models on SNN behavior. Here, we move beyond descriptive use and evaluate how different learning paradigms reshape class-conditional complexity profiles when LZC is used as a decision-relevant temporal descriptor within the classification pipeline. The focus is on characterizing learning-rule-induced changes in temporal organization and relating them to accuracy-efficiency trade-offs under controlled input statistics. 

Table \ref{table:results_SNN_unsupervised} presents the influence of the unsupervised learning algorithm such as Hebbian, STDP, and SDSP on the performance of SNNs. All computations were performed for 10 epochs on an Intel(R) Core(TM) i7-14700F CPU @ 2.10 GHz. A general finding of our study is the pronounced dependence of both classification accuracy and computational overhead on the choice of learning algorithm and the underlying characteristics of the input data. Observed classification accuracies ranged from 89.50\% to 100.00\% across the evaluated methods, whereas computational times exhibited substantial variation, ranging from approximately 7 seconds to over 48,000 seconds.

\par
Table \ref{table:results_SNN_unsupervised} presents the influence of the unsupervised learning algorithm such as Hebbian, STDP, and SDSP on the performance of SNNs. The results obtained for the Hebbian algorithm demonstrate that, within the classical framework, high classification accuracy was achieved for the Bernoulli and Markov processes, with values of 99.00\% and 95.79\%, respectively. These results were accompanied by relatively long computation times of 52.93 s and 74.66 s. Furthermore, MSE values of 0.0011 and 0.00422, suggest minimal prediction errors. The coefficient of determination $R^{2}$ scores of 0.9600 and 0.8313 further confirm a strong agreement between predicted and actual values. For the Poisson process, a classification accuracy of 90.00\% was observed, with a $R^{2}$ score of 0.9074. The corresponding MSE values were 0.1000, while the $R^2$ score of 0.5998 indicates a moderate correlation between the predictions and the actual values. It is worth emphasizing that the accuracy for the Poisson process is achieved for a twice smaller number of neurons in the layers (for Bernoulli and Markov processes, we achieve similar accuracy with the same number of neurons in the layers). An increase in the number of neurons causes a decrease in accuracy below 50.00\%, contrary to the case of Bernoulli and Markov processes, where accuracy increases with increasing number of neurons in layers. Thus, the worse classification results for the Poisson process may be related to the fact that events occur randomly in time, making it difficult to generalize Hebbian learning effectively. The relatively low computation time indicates that the Hebbian learning algorithm efficiently processes structured data. However, taking into account the nature of Hebbian learning it is effective when there are strong, predictable relationships between inputs and outputs, like in the case in Bernoulli or Markov processes. 
\par
The introduction of other bioinspired learning algorithms, namely STDP and SDSP provides a similar level of accuracy, in a similar computational time, with the exception of the SDSP algorithm for Markov processes, where the computational time is an order of magnitude longer. However, the lower MSE for Bernoulli and Markov processes indicate that STDP fine-tunes synaptic weights more effectively than Hebbian learning, reducing classification errors. 
\par
Unsupervised learning algorithms help to achieve high accuracies with much fewer neurons in layers for uncorrelated data.

\begin{table}[H]
\centering
\caption{Influence of unsupervised learning algorithms on SNN performance.}
\label{table:results_SNN_unsupervised}
\footnotesize

\setlength{\tabcolsep}{1.5pt}
\renewcommand{\arraystretch}{0.75}

\begin{tabularx}{\textwidth}{
		>{\raggedright\arraybackslash}p{1.8cm}  
		>{\raggedright\arraybackslash}p{1.5cm}  
		>{\raggedright\arraybackslash}p{1.5cm}  
		>{\centering\arraybackslash}p{1.0cm}    
		>{\centering\arraybackslash}p{1.0cm}    
		>{\centering\arraybackslash}p{0.9cm}    
		S[table-format=3.2]                     
		S[table-format=2.2]                     
		S[table-format=1.4]                     
		S[table-format=1.4]                     
	}
	\toprule
	\textbf{Type} & \textbf{Subtype} & \textbf{Dataset} & \textbf{Bio-} & \textbf{Ep.} & \textbf{$n$} &
	{\textbf{Time [s]}} & {\textbf{Acc. [\%]}} & {\textbf{MSE}} & {\textbf{$R^{2}$}} \\
	\midrule
	
	\multirow{9}{*}{\textbf{Unsupervised}} 
	& \multirow{3}{*}{\textbf{Hebbian}}
	& Bernoulli & Yes & 10 & 128 & 51.93  & 99.00 & 0.0100 & 0.9600 \\
	& & Markov   & Yes & 10 & 128 & 74.66  & 95.78 & 0.0422 & 0.8313 \\
	& & Poisson  & Yes & 10 &  64 & 17.17  & 90.00 & 0.1000 & 0.5998 \\
	\cmidrule(lr){2-10}
	
	& \multirow{3}{*}{\textbf{STDP}}
	& Bernoulli & Yes & 10 &  64 & 62.08  & 99.50 & 0.0050 & 0.9800 \\
	& & Markov   & Yes & 10 & 128 & 513.73 & 98.00 & 0.0200 & 0.9200 \\
	& & Poisson  & Yes & 10 &  64 & 63.36  & 89.50 & 0.1050 & 0.5798 \\
	\cmidrule(lr){2-10}
	
	& \multirow{3}{*}{\textbf{SDSP}}
	& Bernoulli & Yes & 10 & 128 & 43.20  & 98.00 & 0.0200 & 0.9200 \\
	& & Markov   & Yes & 10 & 128 & 46.10  & 93.87 & 0.0613 & 0.7549 \\
	& & Poisson  & Yes & 10 &  64 & 17.83  & 92.50 & 0.0750 & 0.6999 \\
	\bottomrule
\end{tabularx}
\end{table}

Table \ref{table:results_SNN_supervised} presents the influence of the supervised learning algorithms such as backpropagation, tempotron, SpikeProp, Chronotron and ReSuMe on the performance of SNNs. The BP algorithm achieves perfect classification accuracy for Bernoulli, Markov, and Poisson processes, as evidenced by zero MSE, indicating no classification errors. The $R^{2}$ score of 1.000 further confirms its ability to perfectly predict sequence outputs. However, this unmatched accuracy comes at a significant computational cost, with processing times, making BP the most resource-intensive supervised method. 
\par
Although BP is theoretically optimal in terms of predictive performance, its high computational demands render it impractical for real-time or large-scale applications where efficiency is crucial. This trend also holds for the STBP algorithm, but with a significantly greater time expenditure than BP. In turn, in \cite{Yu2014}, the authors state that they studied Bernoulli sources; however, in their network, the input layer consisted of LIF neurons, while the remaining layers comprised perceptrons trained using the tempotron algorithm. They did not present results for this datasets. We achieve high classification accuracy, reaching 99.00\% for Bernoulli process and 100.00\% for Markov process, while in the case of Poisson process the level of accuracy in lower, namely 97.50\%. In the case of tempotron, for Bernoulli and Poisson the calculations are very fast, taking about 7s. For Markov the time is eight times longer, while classification depends not just on the current state but also on prior states, the model must process sequences over multiple time steps, significantly increasing computational demands. Since the Poisson process does not have explicit dependencies like Markov sequences, computations remain efficient.
\par
Also, the computation for the tempotron also requires fewer layers of neurons than BP and STBP with a slight loss of precision. In turn, the SpikeProp algorithm requires fewer neurons in the layers, which affects the computation speed for Markov and Poisson processes. This number decreases for all datasets when we apply the Chronotron learning algorithm, but the computation time is longer only in the case of the Poisson algorithm. Moreover, unlike classical backpropagation, which requires full gradient updates over all layers, Spikprop optimizes synaptic weights through event-driven updates, significantly reducing computational complexity in comparison to BP and STBP.
\par
The ReSuMe demontrate the lowest average classification accuracy among all evaluated supervised learning algotihms. Is quite computationally efficient, while relies on spike pattern association through supervised reinforcement. Consequently, this results in less precise weight corrections, leading to higher classification error. 
\par
Supervised biologically inspired learning algorithms help achieve high accuracies with much fewer neurons in layers for uncorrelated data than supervised algorithms, non-biologically inspired learning algorithms, and unsupervised algorithms.

\begin{table}[H]
\centering
\caption{Influence of supervised learning algorithms on SNN performance.}
\label{table:results_SNN_supervised}
\footnotesize

\setlength{\tabcolsep}{1.5pt}
\renewcommand{\arraystretch}{0.65}

\begin{tabularx}{\textwidth}{
		>{\raggedright\arraybackslash}p{1.8cm}  
		>{\raggedright\arraybackslash}p{1.5cm}  
		>{\raggedright\arraybackslash}p{1.5cm}  
		>{\centering\arraybackslash}p{0.8cm}    
		>{\centering\arraybackslash}p{1.0cm}    
		>{\centering\arraybackslash}p{0.9cm}    
		S[table-format=5.2]                      
		S[table-format=3.2]                      
		S[table-format=1.4]                      
		S[table-format=1.4]                      
	}
	\toprule
	\textbf{Type} & \textbf{Subtype} & \textbf{Dataset} & \textbf{Bio-} & \textbf{Ep.} & \textbf{$n$} &
	{\textbf{Time [s]}} & {\textbf{Acc. [\%]}} & {\textbf{MSE}} & {\textbf{$R^{2}$}} \\
	\midrule
	
	\multirow{42}{*}{\textbf{Supervised}}
	
	& \multirow{8}{*}{\textbf{BP}}
	& Bernoulli      & No  & 10 & 128 &   866.44 &  99.00 & 0.0100 & 0.9600 \\
	&               & Markov         & No  & 10 & 256 &  6211.62 & 100.00 & 0.0000 & 1.0000 \\
	&               & Poisson        & No  & 10 & 512 & 12916.37 &  97.50 & 0.0250 & 0.9000 \\
	&               & MNIST          & No  & 10 & 128 &   207.91 &  99.05 & 0.0094 & 0.9619 \\
	&               & Fashion MNIST  & No  & 10 & 128 &  1106.38 &  95.75 & 0.0425 & 0.8300 \\
	&               & Fashion MNIST  & No  & 10 & 128 &   203.76 &  95.64 & 0.0435 & 0.8257 \\
	&               & N-MNIST        & No  & 10 & 128 &  1508.40 &  99.23 & 0.0072 & 0.9715 \\
	&               & DVS Gesture    & No  & 10 & 128 &   255.77 & 100.00 & 0.0000 & 1.0000 \\
	&               & SHD            & No  & 10 & 128 &   384.71 &  95.26 & 0.0474 & 0.8103 \\
	\cmidrule(lr){2-10}
	
	& \multirow{8}{*}{\textbf{STBP}}
	& Bernoulli      & No  & 10 & 128 &  2261.43 & 100.00 & 0.0000 & 1.0000 \\
	&               & Markov         & No  & 10 & 512 & 48258.21 & 100.00 & 0.0000 & 1.0000 \\
	&               & Poisson        & No  & 10 & 512 & 30093.49 &  96.50 & 0.0350 & 0.8599 \\
	&               & MNIST          & No  & 10 & 128 &   885.10 &  99.83 & 0.0140 & 0.9946 \\
	&               & Fashion MNIST  & No  & 10 & 128 &  1106.38 &  95.75 & 0.0425 & 0.9558 \\
	&               & N-MNIST        & No  & 10 & 128 &  1061.43 &  99.63 & 0.0041 & 0.9837 \\
	&               & DVS Gesture    & No  & 10 & 128 &  1401.89 &  97.92 & 0.0208 & 0.9167 \\
	&               & SHD            & No  & 10 & 128 &  2229.57 &  89.57 & 0.1043 & 0.5270 \\
	\cmidrule(lr){2-10}
	
	& \multirow{8}{*}{\textbf{Tempotron}}
	& Bernoulli      & Yes & 10 & 128 &     7.00 &  96.00 & 0.0400 & 0.8399 \\
	&               & Markov         & Yes & 10 & 128 &    40.50 &  90.59 & 0.0941 & 0.6234 \\
	&               & Poisson        & Yes & 10 & 128 &     7.23 &  89.50 & 0.1050 & 0.5798 \\
	&               & MNIST          & No  & 10 & 128 &    32.20 &  87.55 & 0.1245 & 0.8914 \\
	&               & Fashion MNIST  & No  & 10 & 128 &    38.54 & 100.00 & 0.0000 & 1.0000 \\
	&               & N-MNIST        & No  & 10 & 512 &   593.90 &  98.41 & 0.1776 & 0.2865 \\
	&               & DVS Gesture    & No  & 10 & 128 &   119.15 &  72.95 & 0.2710 & 0.7350 \\
	&               & SHD            & No  & 10 & 128 &    68.71 &  72.04 & 0.2799 & 0.1191 \\
	\cmidrule(lr){2-10}
	
	& \multirow{8}{*}{\textbf{SpikeProp}}
	& Bernoulli      & Yes & 10 & 128 &    64.28 & 100.00 & 0.0000 & 1.0000 \\
	&               & Markov         & Yes & 10 &  64 &    47.00 &  92.34 & 0.0766 & 0.6935 \\
	&               & Poisson        & Yes & 10 &  32 &    20.15 &  91.50 & 0.0850 & 0.6599 \\
	&               & MNIST          & No  & 10 & 128 &    68.80 &  99.12 & 0.0129 & 0.9484 \\
	&               & Fashion MNIST  & No  & 10 & 128 &    92.49 &  93.68 & 0.0632 & 0.7471 \\
	&               & N-MNIST        & No  & 10 & 128 &   383.30 &  98.38 & 0.0162 & 0.9348 \\
	&               & DVS Gesture    & No  & 10 & 128 &   166.38 & 100.00 & 0.0000 & 1.0000 \\
	&               & SHD            & No  & 10 & 128 &    99.37 &  91.94 & 0.0806 & 0.6775 \\
	\cmidrule(lr){2-10}
	
	& \multirow{8}{*}{\textbf{Chronotron}}
	& Bernoulli      & Yes & 10 &  32 &    18.61 &  98.50 & 0.0150 & 0.9400 \\
	&               & Markov         & Yes & 10 &  32 &    46.30 &  92.34 & 0.9300 & 0.0175 \\
	&               & Poisson        & Yes & 10 &  32 &   113.98 &  90.50 & 0.0950 & 0.6198 \\
	&               & MNIST          & No  & 10 & 256 & 16329.10 &  76.18 & 0.2382 & 0.0432 \\
	&               & Fashion MNIST  & No  & 10 & 128 &  2394.12 &  75.25 & 0.2475 & 0.0100 \\
	&               & N-MNIST        & No  & 10 & 256 & 10299.07 &  75.41 & 0.1373 & 0.4482 \\
	&               & DVS Gesture    & No  & 10 & 128 &   197.17 &  87.50 & 0.1250 & 0.8980 \\
	&               & SHD            & No  & 10 &  64 &   100.18 &  81.99 & 0.1801 & 0.2792 \\
	\cmidrule(lr){2-10}
	
	& \multirow{8}{*}{\textbf{ReSuMe}}
	& Bernoulli      & Yes & 10 & 128 &    68.55 &  92.50 & 0.0750 & 0.6999 \\
	&               & Markov         & Yes & 10 &  64 &    46.05 &  91.60 & 0.0840 & 0.6642 \\
	&               & Poisson        & Yes & 10 &  64 &    21.45 &  91.50 & 0.0850 & 0.6590 \\
	&               & MNIST          & No  & 10 & 128 &    56.04 &  98.55 & 0.0145 & 0.9416 \\
	&               & Fashion MNIST  & No  & 10 & 128 &    63.34 &  95.07 & 0.0493 & 0.8029 \\
	&               & N-MNIST        & No  & 10 & 128 &   438.12 &  98.68 & 0.0132 & 0.9470 \\
	&               & DVS Gesture    & No  & 10 & 128 &   171.14 & 100.00 & 0.0000 & 1.0000 \\
	&               & SHD            & No  & 10 & 128 &   101.25 &  91.94 & 0.0806 & 0.6775 \\
	\bottomrule
\end{tabularx}
\end{table}

Table \ref{table:results_SNN_hybrid} presents the influence of hybrid learning algorithms such as ANN-SNN conversion, STPD based on rewards, and BAL on SNN performance. For ANN-SNN conversion, the LZC-based classifier achieves high classification accuracy across all datasets. However, performance for Poisson-distributed inputs remains lower than for Bernoulli and Markov sources due to higher temporal variability. We achieve comparable results with the reward-based and BAL learning algorithms. However in the case of reward based, for Markov and Poisson processes the network requires more neurons in layers than ANN-SNN conversion, but the computation time is shorter. For Bernoulli the network requires twice as few neurons in layers, but the computation time is five times shorter. BAL has six times longer computation time for a Poisson process than the reward based algorithm. Unlike standard Hebbian learning or STDP, BAL seems to deal with stochastic volatility more efficiently.

Thus, Poisson-distributed data presents the greatest challenge but also offers the most biologically realistic testing ground. Across nearly all algorithms, performance was lower for Poisson inputs compared to Bernoulli or Markov, indicating difficulties in processing highly irregular spike trains. It is worth highlighting that biologically inspired and hybrid algorithms like SDSP and BAL demonstrated comparatively better adaptability, underscoring their value for neuromorphic systems. Additionally, the computational burden for Poisson inputs was higher for gradient-based approaches due to lack of structure, which complicates error propagation and increases time complexity.

\begin{table}[H]
\centering
\footnotesize
\caption{Wilcoxon signed-rank test for synthetic stochastic datasets
	(Bernoulli, Markov, Poisson). All comparisons yield $p>0.05$,
	indicating no statistically significant differences between learning rules.}
\label{tab:wilcoxon_synth}

\setlength{\tabcolsep}{2pt}
\renewcommand{\arraystretch}{0.85}

\begin{tabularx}{\textwidth}{
		>{\raggedright\arraybackslash}p{2.5cm}
		X
		>{\centering\arraybackslash}p{2.2cm}
	}
	\toprule
	\textbf{Dataset} & \textbf{Comparison} & \textbf{$p$ (LZC acc.)} \\
	\midrule
	Bernoulli & All algorithm pairs & $>0.05$ \\
	Markov    & All algorithm pairs & $>0.05$ \\
	Poisson   & All algorithm pairs & $>0.05$ \\
	\bottomrule
\end{tabularx}
\end{table}

\begin{table}[H]
\centering
\footnotesize
\caption{Wilcoxon signed-rank test for pairwise comparisons between learning rules
	(Tempotron, SpikeProp, BAL) on neuromorphic datasets. We report $p$-values for
	LZC-based accuracy and raw SNN accuracy. Bold values indicate statistically
	significant differences at $\alpha=0.05$.}
\label{tab:wilcoxon_neuro}

\setlength{\tabcolsep}{2pt}
\renewcommand{\arraystretch}{0.80}

\begin{tabularx}{\textwidth}{
		>{\raggedright\arraybackslash}p{2.6cm}
		>{\raggedright\arraybackslash}X
		>{\centering\arraybackslash}p{2.0cm}
		>{\centering\arraybackslash}p{2.0cm}
	}
	\toprule
	\textbf{Dataset} & \textbf{Comparison} & \textbf{$p$ (LZC acc.)} & \textbf{$p$ (SNN acc.)} \\
	\midrule
	
	MNIST01        & Tempotron vs SpikeProp & 0.32 & 0.32 \\
	MNIST01        & Tempotron vs BAL       & n/a  & n/a  \\
	MNIST01        & SpikeProp vs BAL       & 0.32 & 0.32 \\
	
	\midrule
	FashionMNIST01 & Tempotron vs SpikeProp & 0.11 & 0.07 \\
	FashionMNIST01 & Tempotron vs BAL       & 0.32 & 0.18 \\
	FashionMNIST01 & SpikeProp vs BAL       & 0.10 & 0.07 \\
	
	\midrule
	N-MNIST01      & Tempotron vs SpikeProp & 0.89 & 0.69 \\
	N-MNIST01      & Tempotron vs BAL       & 0.72 & \textbf{0.028} \\
	N-MNIST01      & SpikeProp vs BAL       & 0.83 & \textbf{0.028} \\
	
	\midrule
	Gesture01      & Tempotron vs SpikeProp & 0.18 & 0.59 \\
	Gesture01      & Tempotron vs BAL       & 0.32 & 0.47 \\
	Gesture01      & SpikeProp vs BAL       & 1.00 & 0.27 \\
	
	\midrule
	SHD01 & All pairs & n/a\footnotemark & n/a\footnotemark \\
	
	\bottomrule
\end{tabularx}
\end{table}

Tables~\ref{tab:wilcoxon_synth} and \ref{tab:wilcoxon_neuro} summarize the statistical reliability of LZC-based classification using Wilcoxon signed-rank tests across three SNN training algorithms (Tempotron, SpikeProp, BAL) and multiple data sources. For synthetic stochastic datasets (Bernoulli, Markov, Poisson), all pairwise comparisons yield $p>0.05$, indicating no statistically significant differences between learning rules. A similar trend is observed for neuromorphic datasets, where most comparisons remain statistically  indistinguishable, with only isolated differences appearing in raw SNN accuracy rather than in LZC-based
decisions. These results support the observation that LZC representations produced by different learning rules remain statistically comparable across datasets. Non-significant Wilcoxon results confirm that differences between learning rules are not statistically supported at the level of LZC-based decisions.

\begin{table}[H]
\centering
\caption{Influence of hybrid learning algorithms on SNN performance.}
\label{table:results_SNN_hybrid}

\footnotesize

\setlength{\tabcolsep}{1.5pt}
\renewcommand{\arraystretch}{0.75}

\begin{tabularx}{\textwidth}{
		>{\raggedright\arraybackslash}p{1.8cm}  
		>{\raggedright\arraybackslash}p{1.5cm}  
		>{\raggedright\arraybackslash}p{1.5cm}  
		>{\centering\arraybackslash}p{1.0cm}    
		>{\centering\arraybackslash}p{1.0cm}    
		>{\centering\arraybackslash}p{0.9cm}    
		S[table-format=3.2]                     
		S[table-format=2.2]                     
		S[table-format=1.4]                     
		S[table-format=1.4]                     
	}
	\toprule
	\textbf{Type} & \textbf{Subtype} & \textbf{Dataset} & \textbf{Bio-} & \textbf{Ep.} & \textbf{$n$} &
	{\textbf{Time [s]}} & {\textbf{Acc. [\%]}} & {\textbf{MSE}} & {\textbf{$R^{2}$}} \\
	\midrule
	
	\multirow{9}{*}{\textbf{Hybrid}}
	& \multirow{3}{*}{\textbf{ANN--SNN}}
	& Bernoulli & No  & 10 &  64 & 21.41 & 99.50 & 0.0050 & 0.9800 \\
	&          & Markov    & No  & 10 &  32 & 37.25 & 92.58 & 0.0742 & 0.7031 \\
	&          & Poisson   & No  & 10 &  32 & 115.41 & 94.50 & 0.0550 & 0.7799 \\
	\cmidrule(lr){2-10}
	
	& \multirow{3}{*}{\textbf{Reward}}
	& Bernoulli & No  & 10 &  32 &  4.30 & 99.50 & 0.0050 & 0.9800 \\
	&          & Markov    & No  & 10 &  64 & 36.32 & 90.35 & 0.0965 & 0.9948 \\
	&          & Poisson   & No  & 10 &  64 & 14.83 & 91.00 & 0.0900 & 0.6399 \\
	\cmidrule(lr){2-10}
	
	& \multirow{3}{*}{\textbf{BAL}}
	& Bernoulli & Yes & 10 &  32 &  1.56 & 97.00 & 0.0300 & 0.8800 \\
	&          & Markov    & Yes & 10 & 128 & 46.37 & 95.38 & 0.0462 & 0.8153 \\
	&          & Poisson   & Yes & 10 & 128 & 95.98 & 94.50 & 0.0550 & 0.7799 \\
	\bottomrule
\end{tabularx}
\end{table}

Table \ref{tab:efficiency} provides computational metrics including FLOPs per forward pass, runtime, latency per processed sample, and energy estimates. Energy values are analytical estimates derived from runtime and an assumed average power draw of the measurement platform; they should be interpreted as coarse proxies rather than hardware-measured neuromorphic energy. These results demonstrate that all three SNN learning rules operate in a low-FLOPs regime, with per-sample latency below 0.25 ms for synthetic data and below 0.1 ms for MNIST after optimization, supporting real-time feasibility. It is important to emphasize that Lempel--Ziv complexity is a single-scalar descriptor of temporal structure rather than a discriminative classifier optimized for decision boundaries. When used with a fixed threshold, LZC-based decisions can become unstable under class imbalance or when class-conditional complexity distributions overlap, which may reduce the F1-score even if overall accuracy remains high. In this paper, we therefore interpret LZC primarily as a decision-relevant representation that exposes how learning rules shape temporal information, while standard metrics (accuracy, MSE, $R^2$) quantify separability under the adopted protocol.

\begin{table}[t]
\centering
\caption{Computational efficiency metrics: runtime, latency, FLOPs, and estimated energy per sample.}
\label{tab:efficiency}
\footnotesize

\setlength{\tabcolsep}{1.5pt}
\renewcommand{\arraystretch}{0.75}

\begin{tabularx}{\textwidth}{
		>{\raggedright\arraybackslash}p{1.4cm}
		>{\raggedright\arraybackslash}p{1.2cm}
		S[table-format=2.2]
		S[table-format=2.2(2)]
		S[table-format=1.6]
		S[table-format=1.6]
	}
	\toprule
	\textbf{Dataset} & \textbf{Algorithm}
	& {\textbf{Time per run [s]}}
	& {\textbf{Latency [ms/sample]}}
	& {\textbf{FLOPs [G]/forward}}
	& {\textbf{Energy [mJ/sample]}} \\
	& &
	{\textbf{(mean)}} &
	{\textbf{(mean $\pm$ std)}} &
	{\textbf{(mean)}} &
	{\textbf{(mean)}} \\
	\midrule
	
	Bernoulli & Tempotron & 46.67 & 233.34 \pm 4.17  & 0.000295 & 0.002949 \\
	& SpikeProp & 46.63 & 233.16 \pm 3.74  & 0.000295 & 0.002949 \\
	& BAL       & 23.16 & 115.81 \pm 42.95 & 0.000295 & 0.002949 \\
	\midrule
	Markov    & Tempotron & 46.77 & 233.87 \pm 7.12  & 0.000295 & 0.002949 \\
	& SpikeProp & 47.31 & 236.54 \pm 6.95  & 0.000295 & 0.002949 \\
	& BAL       & 30.75 & 153.77 \pm 65.54 & 0.000295 & 0.002949 \\
	\midrule
	Poisson   & Tempotron & 45.15 & 225.75 \pm 3.76  & 0.000295 & 0.002949 \\
	& SpikeProp & 45.24 & 226.20 \pm 2.48  & 0.000295 & 0.002949 \\
	& BAL       & 28.93 & 144.65 \pm 70.30 & 0.000295 & 0.002949 \\
	\midrule
	MNIST 0/1 & Tempotron & 208.96 & 70.69 \pm 9.00  & 0.000233 & 0.002335 \\
	& SpikeProp & 105.49 & 35.69 \pm 21.89 & 0.000233 & 0.002335 \\
	& BAL       & 71.39  & 24.15 \pm 2.91  & 0.000233 & 0.002335 \\
	\bottomrule
\end{tabularx}
\end{table}

Our experiments reveal that the choice of learning algorithm must be carefully aligned with both the structure of the input data and application-specific computational constraints. Classic backpropagation and surrogate-gradient methods (e.g., STBP) achieve superior accuracy, particularly for structured signals. However, their computational demands are prohibitively high for real-time or resource-constrained deployments. On the other hand, bio-inspired learning algorithms such as tempotron, SDSP, and BAL offer a more favorable trade-off, delivering robust classification with substantially lower computational costs. BAL, in particular, emerges as a promising candidate, balancing biological plausibility with operational efficiency.

Analysis across input types shows that classification accuracy is strongly influenced by the temporal correlation structure of the signals. Poisson-distributed inputs, despite being memoryless and weakly correlated, pose the greatest challenge for learning algorithms due to their high temporal variability and lack of exploitable structure. In contrast, Markovian sequences, with moderate temporal dependencies, offer a balanced scenario, while Bernoulli sequences, being memoryless and maximally entropic per symbol, provide a broad baseline for learning without temporal dependencies. These findings highlight the importance of aligning learning strategies with the statistical properties of input, especially in the design of SNNs for real-world data processing.

In general, achieving brain-like robustness under noisy stochastic input regimes, particularly those modeled by Poisson processes, will require continued refinement of adaptive, efficient, and biologically plausible learning strategies. The LZC-SNN framework proposed here contributes to this goal by offering a scalable, interpretable, and resource-efficient alternative to conventional classification pipelines in spiking neural systems.

While our previous study examined the role of neuron models, the present work demonstrates that learning paradigms exert an equally strong and often dominant influence on the temporal organization of spiking activity, as revealed by complexity-based analysis.

\subsection{Limitations of fixed-threshold LZC decisions}
Lempel-Ziv complexity is a single-scalar summary of temporal novelty and is not optimized to directly maximize discriminative margins. When class priors are imbalanced or when class-conditional LZC distributions overlap, a fixed decision threshold may collapse into predicting the majority class. This leads to deceptively high classification accuracy while driving the F1-score toward zero, indicating poor minority-class recognition.

For this reason, LZC is interpreted in this work primarily as a decision-relevant temporal representation that exposes how learning rules reorganize spike-train structure, rather than as a strong stand-alone classifier. Classification metrics are therefore used as downstream probes of separability under a specified protocol, and class-sensitive measures are reported to explicitly flag regimes in which scalar-complexity decisions become unreliable.

\section{Conclusions}

This study set out to examine the role of learning algorithms in shaping the classification performance and computational efficiency of Spiking Neural Networks. By systematically comparing classical, bioinspired, and hybrid training approaches across datasets with varying temporal structures Bernoulli, Markov, and Poisson processes we were able to characterize how algorithmic choices interact with data statistics to determine both accuracy and runtime.

Our results provide several insights. First, gradient-based methods such as backpropagation and surrogate-gradient learning remain unrivalled in terms of predictive precision, consistently achieving near-perfect accuracy and negligible error values. However, their computational costs were prohibitively high, in some cases exceeding 48,000 seconds (measured over 10 epochs on a standard desktop machine with Intel® Core™ i7-14700F @ 2.10 GHz), which makes them impractical for real-time or embedded neuromorphic deployments. Second, bioinspired algorithms such as Hebbian learning, STDP, SDSP, tempotron, and SpikProp offered substantially lower computational costs, often two orders of magnitude faster, while still maintaining competitive accuracy for structured sequences. Their limitations became apparent in the context of Poisson-distributed data, where stochastic variability hindered generalization and accuracy degraded. Third, hybrid schemes including ANN-SNN conversion and BAL were shown to mitigate some of these weaknesses by leveraging biologically plausible mechanisms that adapt more flexibly to irregular temporal inputs.

The methodological novelty of this work lies in the use of Lempel-Ziv complexity as a decision-relevant temporal descriptor embedded within an SNN framework. Unlike prior studies where LZC served primarily as a descriptive post hoc metric, here it is used to analyze how different learning rules shape temporal representations and influence classification-related outcomes, rather than as a standalone discriminative classifier. This integration of information-theoretic complexity analysis with biologically plausible spiking computation advances the field in two important ways: it improves interpretability by linking decision boundaries to structural novelty in the input, and it enhances efficiency by reducing reliance on gradient-based training. We believe this conceptual bridge between complexity theory and neuromorphic computing can serve as a useful template for future research in pattern recognition with SNNs.

At the same time, several limitations must be acknowledged. The experiments were conducted on synthetic datasets, which, while useful for systematically controlling temporal correlations, may not capture the full irregularities of real-world neural or sensory data. Moreover, although LZC-SNN achieved competitive accuracy, the method requires further validation on larger-scale problems and with hardware implementations to fully assess scalability. Finally, while we report coarse energy estimates, hardware-measured energy on neuromorphic processors remains out of scope and will be addressed in future work.

Despite these limitations, the results carry implications for both researchers and practitioners. For researchers, the comparative framework presented here provides a benchmark for assessing novel SNN learning rules under varying temporal conditions. For practitioners designing neuromorphic systems, the findings clarify trade-offs between accuracy and efficiency and highlight which algorithms are best suited for structured versus stochastic inputs.

The main strength of our approach lies in its interpretability efficiency compared to conventional decoding strategies. At the same time, its accuracy is sensitive to the statistical properties of the input and to the choice of learning rules, which we identified as a limiting factor for real-world deployment. These insights can guide researchers in selecting appropriate training strategies depending on task constraints.

Future work should proceed along several directions. Extending the evaluation to real-world datasets such as electrophysiological recordings, speech signals, or event-based vision would test the robustness of the observed trends. Hardware level experiments are needed to quantify energy savings and latency reductions in neuromorphic processors. Methodologically, there is an opportunity to design adaptive or hybrid algorithms that dynamically switch between gradient-based and bioinspired updates depending on the input statistics, thereby combining accuracy with efficiency. Finally, further exploration of information-theoretic measures beyond LZC, such as entropy rate or predictive information, may provide complementary tools for interpretable spiking computation.

In summary, this work highlights the decisive role of learning algorithms in determining the balance between accuracy, efficiency, and biological plausibility in SNNs. By embedding complexity-theoretic principles into spiking architectures, we take a step toward interpretable and resource-efficient pattern recognition systems that more closely align with the constraints of real-world neuromorphic applications.

\section{Practical Implications}

The findings suggest that classical gradient-based algorithms should be reserved for offline or high-resource scenarios requiring maximal accuracy, whereas bioinspired and hybrid methods provide a more favorable balance for real-time and resource constrained neuromorphic applications. The proposed LZC-SNN framework, by combining spiking dynamics with information-theoretic analysis, offers a scalable and interpretable approach to temporal pattern recognition, with potential applications in brain-computer interfaces, biomedical signal analysis, and event-driven sensory processing.

\end{document}